# Characterizing Disparity Between Edge Models and High-Accuracy Base Models for Vision Tasks

Zhenyu Wang
University of North Carolina at Chapel Hill
Chapel Hill, USA
zywang@cs.unc.edu

Shahriar Nirjon
University of North Carolina at Chapel Hill
Chapel Hill, USA
nirjon@cs.unc.edu

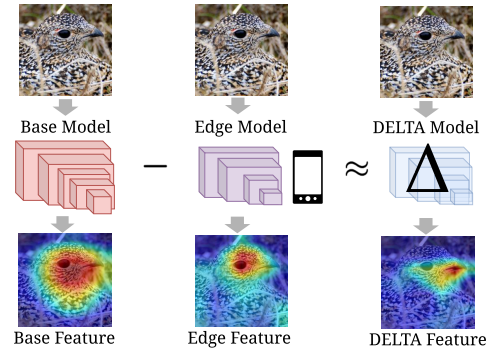

Figure 1: XDELTA provides explanations for why the *edge* model predicts incorrectly, contrasting with the accurate predictions of the *base* model by leveraging the proposed DELTA network's feature representation.

## ABSTRACT

Edge devices, with their widely varying capabilities, support a diverse range of edge AI models. This raises the question: *how does an edge model differ from a high-accuracy (base) model for the same task?* We introduce XDELTA, a novel explainable AI tool that explains differences between a high-accuracy *base* model and a computationally efficient but lower-accuracy *edge* model. To achieve this, we propose a learning-based approach to characterize the model difference, named the DELTA network, which complements the feature representation capability of the *edge* network in a compact form. To construct DELTA, we propose a sparsity optimization framework that extracts the essence of the *base* model to ensure compactness and sufficient feature representation capability of DELTA, and implement a negative correlation learning approach to ensure it complements the *edge* model. We conduct a comprehensive evaluation to test XDELTA's ability to explain model discrepancies, using over 1.2 million images and 24 models, and assessing real-world deployments with six participants. XDELTA excels in explaining differences between *base* and *edge* models (arbitrary pairs as well as compressed *base* models) through geometric and concept-level analysis, proving effective in real-world applications.

## 1 INTRODUCTION

Recent literature has proposed numerous neural network models for edge devices that aim to solve the same learning task, e.g., image classification and object detection. These models exhibit variations in the number of parameters, computational costs, learning capabilities, and overheads—due to the diverse range of edge platforms they run on [52, 59]. While some of these models are developed from scratch [12, 32], many are derived from large and complex models [5, 49] through various compression and transformation techniques [33, 34] to satisfy the resource limitations of edge devices. Even after deployments, these models continue to evolve as they are trained on new data [18, 42], and/or when their architectures are modified [4, 38]. With so many model variants, it becomes crucial [37] to understand the differences between their decision-making processes, which offers more insight than mere comparison of accuracy numbers.

Unfortunately, existing explainable AI techniques that are primarily developed for standalone models [6, 45] are inadequate for comparing and contrasting the capabilities and differences between a high-accuracy *base* model and a computationally efficient but lower-accuracy *edge* model. Firstly, they do not provide any relative explanations. Their focus is solely on explaining individual model behavior on specific inputs, which does not provide interpretable relative differences between a pair of models. Secondly, they do not provide generalizable explanations. Their outcome is too specific to the given input, which does not generalize across many examples and multiple datasets. Recent works on model similarity analysis enable architectural-wise segment equivalence measurement and distance-based similarity metric comparison [11, 16, 24]. However, these techniques fall short in effectively explaining the fine-grained specifics of differences in instance-based decision-making processes among various *base* and *edge* models. Hence, devising a new technique that characterizes the generatlized relative difference between a *base* and an *edge* model has remained an open problem.

In this paper, we propose — XDELTA, a new kind of explainable AI technique that categorizes and summarizes the explanations behind an *edge* model's relatively poor performance compared to a *base* model. It performs both geometric and semantic explanation analysis of the disparities between the feature maps of two models and provides a breakdown of cases where the *edge* model fails to correctly classify but the *base* model succeeds. The development of XDELTA involves two major tasks: (1) construction of a difference or DELTA model that represents the relative difference between *base* model and *edge* model, and (2) the generation and summarization of explainable differences between the model pair.

The DELTA model plays a critical role in XDELTA. Figure 1 illustrates how the DELTA model behaves in XDELTA. Given a high-accuracy *base* model and a relatively lower-accuracy *edge* model, we construct a DELTA model that is *compact* in its size and complexity, and is *complementary* in its ability to enhance and rectify the feature representation of the *edge* model, as such, when the *edge* model and the DELTA model are fused, their combined performance is akin to the *base* model. We note that the goal of DELTA is not to approximate the *base* model; rather, it aims to approximate the difference between the *base* and the *edge* model. This is illustrated by the activation maps in Figure 1.

The construction of DELTA is a challenging feat. First, since the DELTA model complements the *edge* model, it primarily needs to capture the essence of the *base* model in a compact network architecture with reduced model complexity and high efficiency (adhering to Occam's Razor principle). We formulate this as a subgraph extraction problem, where the objective of the subgraph is to preserve the *base* model's partial feature-representation capabilities, rather than to maintain the accuracy of the *base* model in its entirety—since the *edge* model also contributes its features when DELTA is fused with it. Second, the DELTA model needs to extract those features from the input that complement the *edge* model's feature representations. This requires a different kind of algorithmic design that not only considers the overall representation capability of the fused model but also carefully make the contributions of the *edge* and DELTA models complementary.

To address these challenges, we devise a new structured subgraph extraction algorithm that is well-suited to DELTA models and a new objective function that considers both the feature representation quality of the fused model while keeping the feature maps of the DELTA and the *edge* models negatively correlated (i.e., complementary) to each other. Once the DELTA model is obtained, it is applied to the test dataset, which may include previously unseen images, to infer and analyze the semantic concepts missed by the *edge* model but captured by the *base* model. This helps understand the reasons behind the *edge* model's subpar performance at the human-understandable semantic level.

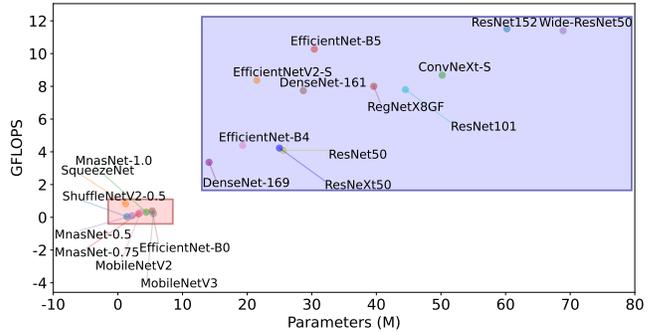

Figure 2: Parameter and FLOPS distribution of various CNN models for ImageNet-1K [43] classification task.

We extensively evaluate XDELTA's efficacy and algorithmic aspects across various scales and configurations. We leverage four popular image datasets containing over 1.2 million images and 24 models derived from 11 image classifiers. Additionally, 6 participants assess explanations in a real-world indoor scene recognition deployment. We demonstrate that XDELTA excels at explaining differences between *base* and *edge* models from multiple perspectives. It employs high-level geometric categorization to quantify activation region patterns, revealing deficiencies in the *edge* model. Furthermore, XDELTA provides fine-grained concept-level explanations, identifying missing semantic concepts during *edge* model decision-making and explaining misclassifications. Notably, XDELTA performs well with both arbitrary model pairs and compressed models (*edge* versions of *base* models). Finally, the real-world deployment demonstrates XDELTA's ability to explain differences between *edge* and *base* models across eight environment categories, using a total of 421 mobile phone images.

## 2 MOTIVATION

A wide variety of neural networks with different parameter sizes, computational costs, learning capabilities, and overheads exist that solve the same learning task. In Figure 2, for example, we plot 20 popular CNNs for the image classification task. Models in the lower-left rectangle have lower parameter sizes and computational costs (FLOPS) compared to those in the upper-right rectangle. These smaller, computationally efficient models, aka the *edge* models, are generally suitable for resource-constrained edge devices but usually exhibit lower accuracy than state-of-the-art *base* models designed for high-end machines. We aim at understanding the reasons behind this accuracy gap between an *edge* model and a high-accuracy *base* model.

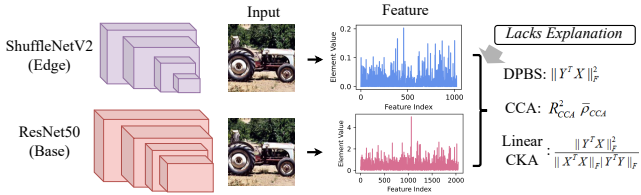

Figure 3: Analyzing model difference with features.

**Analyzing Model Difference Using Performance Metrics.** One straightforward way to compare two models is to directly compare their performance metrics, such as parameter size, FLOPS, and accuracy. For example, ShuffleNetV2, in comparison with ResNet50, is 18 times smaller, 100 times less computationally intensive, and 15% less accurate on the ImageNet-1K dataset. While these performance metrics are simple, they only provide numerical values without explaining the reasons behind the accuracy loss.

**Analyzing Model Difference Using Feature Similarity.** Another, more advanced approach for analyzing model differences is to compare feature representations. As shown in Figure 3, features extracted from the last layer of each model's feature extractor are analyzed for similarity or correlation using various metrics such as Canonical Correlation Analysis [41], Linear Centered Kernel Alignment [17], and Dot Product-Based Similarity [17]. This method is superior to simply comparing accuracy or parameters because it provides insights into how models represent data internally, revealing more about their learning processes and potential strengths or weaknesses. However, these metrics still provide numerical values representing latent feature space information that is difficult to interpret and lacks a direct explanation of the model's decision-making behavior.

**Analyzing Model Difference Using Activation Map.** The *class activation map* has been shown in recent literature to be effective in explaining the decision-making process of a neural network. This map highlights the important regions of the input that influence the inference decision [58]. Based on this, in Figure 4 (a), we show a straightforward approach to compare two models which involves identifying and analyzing the regions where the two activation maps differ. Metrics like the Jaccard Index [55], Dice Coefficient [47], and Overlap Coefficient [51] are used for this analysis. While this method helps visualize the difference between two models for a given input, it is too specific to individual input images as shown in Figure 4 (b). It does not generalize across examples or datasets. It also fails to fully describe fundamental differences in model representations, especially when non-linear processes are involved during map creation [45]. We make two observations from these maps:

- Activation maps vary significantly even if two classifiers make the same prediction. Notably, for correct predictions,

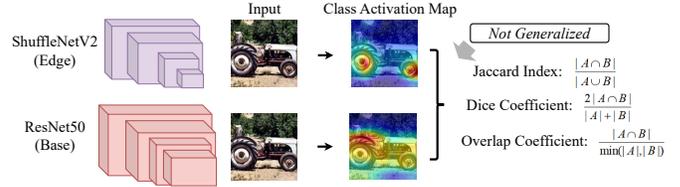

(a) Analyzing model difference with class activation maps.

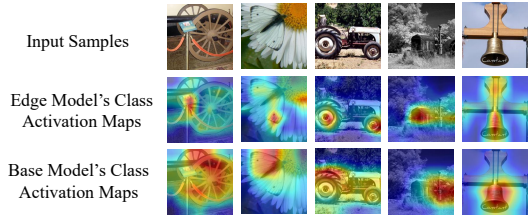

(b) Varying differences in class activation maps of input samples that are correctly predicted by *base* model but misclassified by *edge* model.

Figure 4: Model differences with class activation maps.

this means there is no unique correct way to represent an object. Therefore, why an *edge* model mispredicts should be discussed relative to a specific *base* model.

- Mispredictions are caused by a variety of activation map mismatches such as – completely disjoint, or partially overlapped regions in the foreground and/or the background of the input. The semantics of these regions vary tremendously across different inputs as well. Hence, it requires complex modeling and many examples to characterize or generalize why an *edge* classifier mispredicts.

These observations motivate us to design an algorithm that learns the complex pattern of activation map mismatches and their impact on neural network decisions. The algorithm produces a compact, complementary model representing the difference or *delta* between two models. This model serves as a diagnostic tool to explain performance differences and as a repair tool to enhance a lower-accuracy *edge* network.

## 3 OVERVIEW OF XDELTA

XDELTA is an explainable AI technique that categorizes and summarizes the explanations behind a computationally efficient but relatively lower-accuracy *edge* model's relatively poor performance compared to a high-accuracy *base* model[1]. The development of XDELTA involves two major tasks: (1) the construction of a model that characterizes the difference between the model pair, and (2) the generation and summarization of explainable semantic differences between the model pair. In this paper, we limit our scope to CNNs for image classification tasks considering their prevalence in recent embedded AI literature [35, 39]. For other types of models, the framework to implement XDELTA will be the same, but the details will be slightly different.

---

[1]The model pair is defined as having been fully trained and exhibiting a minimum accuracy difference (of certain percentage) on a test dataset. The models we study in this paper, have at least 10% accuracy differences.

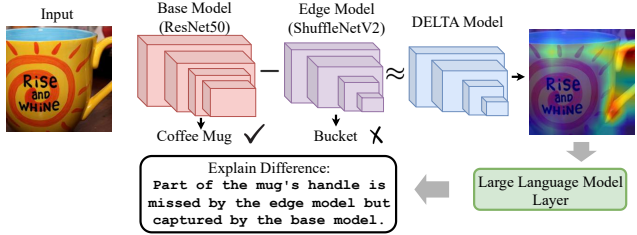

Figure 5: XDELTA explains the essential information missed by *edge* model that leads to a wrong prediction.

## 3.1 DELTA Network Construction

Given a compatible model pair, we define DELTA as a network that models the shortcomings in the *edge* model's data representation ability, relative to the *base* model's ability to do the same. A DELTA network has two salient properties:

- The DELTA model should complement the *edge* model's feature representation such that when the two models are fused together, the accuracy of the fused model is as high as the *base* model's accuracy.
- The DELTA network should be minimal in size and execution cost (FLOPS) to adhere to the Occam's Razor principle. As such, the combined size as well as the execution cost of the DELTA and the *edge* model should be lower than the *base* model's size and execution cost, respectively.

Two major challenges lie ahead in constructing DELTA network: (1) devising a network that preserves partial knowledge from *base* model, such that it complements features of *edge* model; (2) ensuring the compactness and efficiency of DELTA model while retaining the capability of capturing the features missed by *edge* model.

## 3.2 Differential Explainable AI

Given a constructed DELTA network, several steps are required to clearly illustrate the differences between the corresponding model pair. We employ an existing explainable AI tool to generate a class activation map for the input, specifically highlighting areas captured by the *base* model but missed by the *edge* model. This map is then fed into the Large Language Model (LLM) layer to generate detailed and readable explanations, illustrating the differences. Additionally, we provide a geometric and semantic categorization summary outlining the behavior of the DELTA model, which emphasizes differences between the *base* model and various types of *edge* models (either directly derived from the *base* model or completely different ones).

Figure 5 illustrates the concept of XDELTA. When given an input image of a coffee mug, the *edge* model misclassifies it due to its inability to capture information from the handle, as observed in the class activation map generated by the DELTA model. This results in a prediction of a bucket, focusing primarily on the mug's body. The generated map is then passed to the LLM layer, which produces readable concept-level explanations highlighting the model difference.

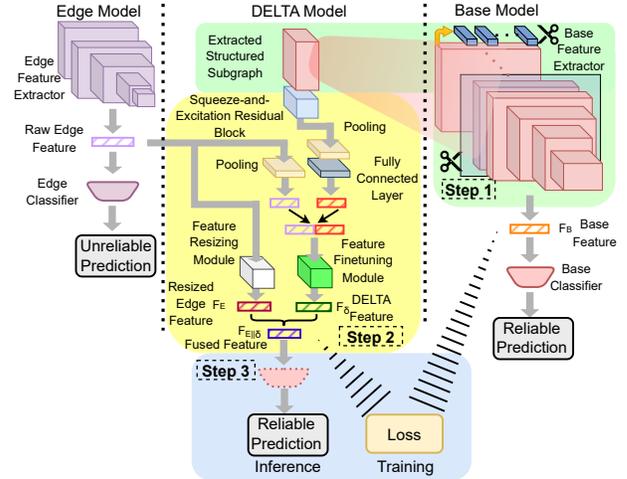

Figure 6: DELTA model architecture.

## 4 DELTA NETWORK CONSTRUCTION

### 4.1 Overview

A DELTA model enhances the data representation capability of an *edge* network in a compact and complimentary manner as such when it's fused with the *edge* network, their combined data representation capability becomes as close as possible to the *base* model's. To achieve this, we propose a network architecture for DELTA that resembles the English letter $Y$ where the two branches capture the essence of the *base* and the *edge* model, respectively, and information flowing through these two branches are fused by the remaining part of the network to construct a representation of the input that we call the DELTA feature. During inference, DELTA features are used in conjunction with features extracted by the *edge* network to classify the input.

The architecture of a DELTA network along with the corresponding *base* and the *edge* network architectures is shown in Figure 6. The construction of DELTA has three steps:

- **Step 1** – Extracting structured subgraph from the *base* network to construct a portion of the DELTA that brings the *base* network's data representation capability into it.
- **Step 2** – Constructing the DELTA feature extractor that adapts and concatenates features extracted by the *edge* model and the subgraph of the *base* model.
- **Step 3** – Training the DELTA network by explicitly ensuring that the DELTA network is complementary to the *edge* network's ability to classify input data.

## 4.2 Step 1 – Structured Subgraph Extraction

In this step, a major part of the DELTA network is constructed by extracting the structured subgraph from the *base* network. The goal of subgraph extraction is not to have a smaller model that performs as accurately as the *base* model, but to extract feature representation capability from the *base* model, which when combined with the *edge* model's feature representation capability, the combined model performs as accurately as the *base* model. We describe the fundamentals and rationale behind developing a new subgraph extraction technique, followed by an optimization framework to obtain the parameters for the extraction process, and finally, the structured subgraph extraction steps.

**Fundamentals and Rationale.** Structured subgraph extraction entails application of binary masks to network weight matrices in order to reduce their size and execution cost:

$$W_{\text{sparse}} = M(\varsigma) \otimes W_{\text{dense}} \quad (1)$$

where, $W$ and $M(\varsigma)$ are the weight and the mask matrices; $\varsigma$ denotes the sparsity of the mask (i.e., number of zero elements divided by total number of elements); $\otimes$ is the element-wise product operation.

A large body of recent works are dedicated to finding optimal sparsity rates for different network constructs [26, 49]. A fundamental limitation of these works, however, is that the search space being too large, they get stuck in a bad local minima and perform sub-optimally. We observe that instead of directly searching for sparsity rates, if we redefine sparsity as a convex combination of *n* candidate sparsity rates $\{\varsigma_i\}$ with the corresponding sparsity coefficients $\{\gamma_i\}$, the search converges fast and becomes resilient to getting stuck in a bad local minima:

$$\varsigma = [\gamma_1\ \gamma_2\ \cdots\ \gamma_n] \times [\varsigma_1\ \varsigma_2\ \cdots\ \varsigma_n]^T \quad (2)$$

The mathematical insight behind the above observation is that when the loss function $\mathcal{L}(W)$ reaches a local minima for weight $W_0$, $\mathcal{L}(W)$ is convex with respect to $W \in [W_0 - \epsilon, W_0 + \epsilon]$, where $\epsilon$ is a small perturbation. Hence, when a mask $M(\varsigma)$ is applied, the loss function $\mathcal{L}(W \otimes M(\varsigma))$ becomes convex with respect to $M(\varsigma)$ given a frozen $W$. To solve this convex optimization problem, we assume $M(\varsigma) \in [0, 1]^n$ is continuous (and discretize afterwards without violating correctness). Using Jensen's inequality:

$$\mathcal{L}\left(W \otimes \sum_{i=1}^{n} \gamma_i M(\varsigma_i)\right) \leq \sum_{i=1}^{n} \gamma_i \mathcal{L}\left(W \otimes M(\varsigma_i)\right) \quad (3)$$

Based on the above, we can always find a subset $\{\hat{M}(\varsigma_j)\}$ for which the following holds:

$$\mathcal{L}\left(W \otimes \sum_{i=1}^{n} \gamma_i M(\varsigma_i)\right) \leq \min_{j} \mathcal{L}\left(W \otimes \hat{M}(\varsigma_j)\right) \quad (4)$$

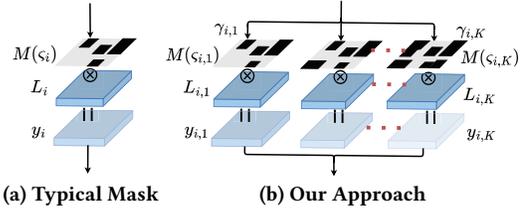

(a) Typical Mask     (b) Our Approach

Figure 7: Illustration of extracting subgraph from one layer.

PROOF. To prove Equation (4) by contradiction, we first simplify it by setting: $W \otimes \sum_{i=1}^{n} \gamma_i M(\varsigma_i) \to X$, $W \otimes M(\varsigma_i) \to Y_i$, and then assume it's incorrect, i.e., $\forall i, \mathcal{L}(X) > \mathcal{L}(Y_i)$. If so, then $\sum_{i=1}^{n} \gamma_i \mathcal{L}(Y_i) < \sum_{i=1}^{n} \gamma_i \mathcal{L}(X) = \mathcal{L}(X) \sum_{i=1}^{n} \gamma_i = \mathcal{L}(X)$, which violates (4). Hence, Equation (4) holds. □

In other words, Equation (4) shows that a mask formed by weighted averaging keeps the loss at a relatively lower value. Since the loss remains closer to the local minima, the search process takes less time to converge and accuracy preservation becomes easier.

**Optimization Framework.** Given the new definition of sparsity, a new challenge is to find the optimal sparsity coefficients $[\gamma_1\ \gamma_2\ \cdots\ \gamma_n]$ for each layer that is processed. These coefficients are used to generate masks that are applied to corresponding layers for subgraph extraction. The optimization goal is to minimize the loss of essential feature representation capability as well as the execution cost of the extracted subgraph. The entire framework runs on high-end server to expedite the optimization process.

To achieve this, an extended version of the *base* network that explicitly incorporates sparsity coefficients is constructed. For each convolutional and linear layer $L_i$, $K$ copies are created, and for each copy $L_{i,j}$, a sparsity rate $\varsigma_{i,j} \in \{\varsigma_1, \cdots, \varsigma_n\}$ and a learnable weight $\gamma_{i,j}$ is assigned. A larger $K$ allows finer search space exploration for optimal solutions but is limited by the server's computational capacity.

Figure 7 shows different approaches to extract subgraph from one layer. The same applies to all layers being processed. The output $y_i$ of layer $L_i$ is the weighted combination of the outputs of all $L_{i,j}$ branches: $y_i = \Sigma_{j=1}^{K} |\gamma_{i,j}| \times y_{i,j}$. The loss function of the network is as follows:

$$\mathcal{L}_p = \lambda_0 \mathcal{L}_{CE} + \lambda_1 \sum_{i=1}^{L} \sum_{j=1}^{K} |\gamma_{i,j}|\ (\alpha \cdot G_{i,j} + \beta \cdot H_{i,j}) \quad (5)$$

where $\mathcal{L}_{CE}$ is the cross-entropy loss that accounts for accuracy to measure the quality of represented feature. The second term accounts for the cost of memory access $G_{i,j}$ and the number of multiply-accumulate operations per second $H_{i,j}$. Incorporating these costs into the loss function yields computationally inexpensive substructures suitable for resource-constrained systems. $\lambda_0$, $\lambda_1$, $\alpha$ and $\beta$ are hyperparameters controlling the relative strength of each term.

During the training of the extended network, masked parameters are kept frozen as such no gradient flows through them. Once the training process converges, the sparsity coefficients are recorded, which are used in the subgraph extraction process. The extended network is discarded at this point since it is no longer needed.

**Subgraph Extraction Process.** Once the optimal sparsity coefficients $\{\gamma_{i,j}\}$ are obtained, we take their absolute values and normalize as $0 \leq \gamma_{i,j} \leq 1$, $\Sigma_{j=1}^{K}\gamma_{i,j} = 1$, the sparsity rate $\varsigma$ is computed using Equation (2). Corresponding masks $M(\varsigma)$ are generated by applying a $l_2$ norm-based importance ranking technique [9, 22] and then applied to each layer of the *base* model that is to be extracted using Equation (1). This results in a subgraph of the *base* model which is fine-tuned further by retraining. The fine-tuning process includes model scaling [13, 50] and several data augmentation methods such as random cropping [3], horizontal flipping [3], random perspective adjustment [3], color jittering [3], and label smoothing [36].

The obtained subgraph is compact, yet its size is further reduced by removing the last few layers, since they incur high overhead but offer limited semantic gain. This results in an extremely compact DELTA model that adheres to size and FLOPS constraints. While dropping these layers diminishes the subgraph's representational capability, it still effectively complements the *edge* model and fulfills its intended purpose.

### 4.3 Step 2 – Constructing DELTA Feature

In this step, features from the *edge* and the subgraph extracted from *base* networks are fine-tuned and combined to form the DELTA features.

**Fine-Tuning Features.** In order to improve the representational capability of the *base* network's subgraph, a squeeze-and-excitation [14] block with *skip connection* [12] is added to obtain the interdependencies between feature channels. These are further passed through a *global average pooling layer* [25] to ensure that the number of elements in the feature vector is the same as the channel size of the input feature map. Finally, the features are reshaped and linearly transformed to a lower dimension using a fully-connected layer. The *edge* model is unchanged, and the features from the *edge* network is only down-scaled by a pooling layer to incorporate enough information that helps speedup the convergence of DELTA feature construction process.

**Combining Features.** The fine-tuned features are concatenated to form an extended feature vector – which is fed to a two-layer perceptron (MLP) network that acts as a feature resizer. This resizer is essential for merging features with different dimensions. It ensures the combined feature after resizing can be averaged with the down-scaled features from the *edge* model and then fed to the *base* model's classifier (i.e., layers after the feature extractor) for comparable accuracy.

### 4.4 Step 3 – Training DELTA Network

In this step, the DELTA network is trained while the *edge* and the *base* models remain frozen. Considering the fused feature quality, complementary nature, and efficiency, the loss function includes three terms — mean squared error ($\mathcal{L}_{MSE}$), feature-wise negative correlation ($\mathcal{L}_{FNC}$), and sparsity regularization ($\mathcal{L}_{SR}$). The hyperparamters $\lambda_{FNC}$ and $\lambda_{SR}$ control the relative strength of corresponding terms.

$$\mathcal{L} = \mathcal{L}_{MSE} + \lambda_{FNC} \times \mathcal{L}_{FNC} + \lambda_{SR} \times \mathcal{L}_{SR} \quad (6)$$

**Mean Squared Error (MSE).** This term ensures that when the DELTA features are fused with the *edge* model, their combined accuracy is on par with the *base* model's accuracy.

$$\mathcal{L}_{MSE} = \frac{1}{N} \sum_{i=1}^{N} \left( F_{E||\delta}^{(i)} - F_B^{(i)} \right)^2 \quad (7)$$

where $F_{E||\delta}^{(i)}$ and $F_B^{(i)}$ are feature representations of the fused model and the *base* model for the *i*-th ($1 \leq i \leq N$) training example.

**Feature-wise Negative Correlation (FNC).** This term ensures that DELTA is complementary to the *edge* model as such their features are negatively correlated to each other. It forces DELTA to learn different regions on the activation map than what the *edge* model attends to while the fused feature representation is as close to the *base* model's as possible. Unlike traditional negative correlation learning approaches [30] that penalize two models at the instance level, we introduce a feature-wise correlation loss designed for DELTA models:

$$\mathcal{L}_{FNC} = \frac{2\lambda}{N} \sum_{i=1}^{N} \left( \left( F_E^{(i)} - F_{E||\delta}^{(i)} \right) \left( F_\delta^{(i)} - F_{E||\delta}^{(i)} \right) \right) \\ + \frac{1}{2N} \sum_{i=1}^{N} \left( \left( F_\delta^{(i)} - F_B^{(i)} \right)^2 + \left( F_E^{(i)} - F_B^{(i)} \right)^2 \right) \quad (8)$$

where $F_\delta^{(i)}$ and $F_E^{(i)}$ denote the DELTA feature and resized *edge* feature for the *i*-th ($1 \leq i \leq N$) training example, respectively, and $\lambda$ is a hyperparameter that adjusts the strength of the correlation penalty.

**Sparsity Regularization (SR).** The sparsity of the DELTA network is regularized by penalizing the absolute magnitude of its weights so that the model retains only the relevant features.

$$\mathcal{L}_{SR} = \sum_{l=1}^{\#\text{layers}} \left( \sum_{\{f\}} \|W_l^f\|_g + \sum_{\{c\}} \|W_l^c\|_g \right) \quad (9)$$

where $\|W_l^f\|_g$ and $\|W_l^c\|_g$ denote the group lasso [53] for filter- and channel-wise weights for *l*-th convolutional layer.

## 5 DIFFERENTIAL EXPLAINABLE AI

XDELTA analyzes the class activation map of the DELTA network to generate a summary of the *edge* model's shortcomings that is generalizable across multiple datasets. It leverages the DELTA model to summarize cases where an *edge* model underperforms compared to a *base* model on a given dataset. There are three simple steps:

- **Step 1 – Initialization:** An existing explainable AI tool, e.g., GradCAM++ [6] is used to obtain the activation map of the DELTA network for each image in the dataset. These activation maps are segmented (by applying a threshold on activation values) to obtain one or more disjoint activation regions. This process is repeated for the *edge* model.
- **Step 2 – Geometric Categorization:** For each image, activation regions from the *edge* and the DELTA are geometrically compared and categorized into one of seven predefined categories. A summary statistics is produced by counting the occurrences for each category. Figure 8 shows an example.
- **Step 3 – Semantic Categorization:** This step applies when the input dataset contains fine-grained semantic labels for different segments inside each image. The activation regions of the DELTA model inherit those semantic labels depending on their overlaps with the labeled segments. In cases where ground truth semantic labels are unavailable, we employ the GPT-4o LLM to generate the labels. A summary is produced by counting the occurrences of each semantic label. Figure 9 shows an example.

## 6 DATASET-DRIVEN EVALUATION

### 6.1 Dataset and Models

We conduct experiments on four image datasets: ImageNet-1K [43], CIFAR10 [19], MIT Indoor Scenes [40], COCO [28]. For each dataset, we choose a representative pair of models whose input dimensions are compatible with the dataset, as shown in Table 1. We report geometric summaries for the first three datasets and a semantic summary for the last one.

To study the effect of compression on lost semantics, we use ResNet56 [12] as the *base* model pre-trained [1] on CIFAR10 dataset. A fine-tuning dataset is constructed using the ImageNet samples based on the records of the enlarged CINIC10 [7]. This fine-tuning dataset contains similar categories as CIFAR10 dataset while preserving higher resolution and more details. We use a randomly generated mask with controlled global sparsity to perform structured pruning on the ResNet56 using $l_2$-norm as pruning criteria. As a result, we obtain four models: a *base* model and three edge models *edge0*, *edge1* and *edge2*, as shown in Table 2.

| Dataset | Base Model | Edge Model | Summary Type |
|---|---|---|---|
| ImageNet-1K | ResNet50 | ShuffleNetV2 | Geometric |
| CIFAR10 | VGG16 | AlexNetS | Geometric |
| MIT Indoor Scenes | ResNet18 | SqueezeNet | Geometric |
| COCO | ResNet50 | ShuffleNetV2 | Semantic |

Table 1: Models and dataset

| Models | base | edge 0 | edge 1 | edge 2 |
|---|---|---|---|---|
| Compression Ratio (%) | 0 | 89.15 | 97.79 | 98.58 |
| Accuracy (%) | 80.10 | 69.12 | 61.07 | 53.80 |

Table 2: Configuration of *base* model and *edge* models

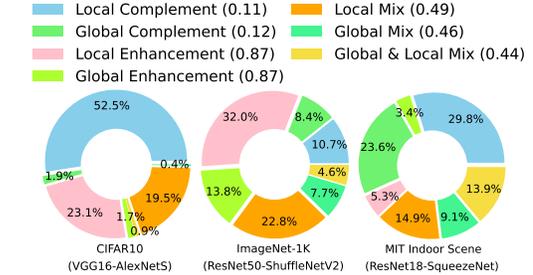

(a) Summary of geometrically categorized explanations.

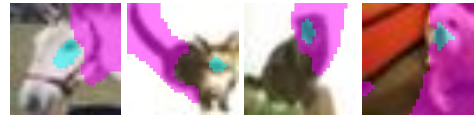

Local Complement      Local Enhancement

(b) Examples of geometric categorization from CIFAR10.

Figure 8: Geometrically categorized explanations of different model pairs and datasets.

### 6.2 Geometric Categorization

Activation regions of the *edge* and the DELTA models show different degrees of overlap — having different impacts on the *edge* model. Disjoint regions *complement* the *edge* model by bringing missing information. Overlapping regions *enhance* the *edge* model by suppressing noise. Often complementing and enhancing regions appear in a *mix*. These regions may be located on or near *local* segments of the target object, or spread out *globally* on other parts of the image.

Figure 8 (a) explains the incorrect predictions by the *edge* models by categorizing the reasons into seven different categories. For the VGG16-AlexNetS pair on CIFAR10, local complementary regions explain the majority of its mispredictions since AlexNetS misses a large number of important regions on the target object. For the ResNet50-ShuffleNetV2 pair on ImageNet-1K, the local enhancement explains the majority of its mispredictions since ShuffleNetV2 attends to many noisy regions which are suppressed by the DELTA model. For the ResNet18-SqueezeNet pair on MIT Indoor Scenes dataset, both global spatial features and local object features are complemented by the DELTA model to correct most of the mispredictions by the SqueezeNet. We also quantify

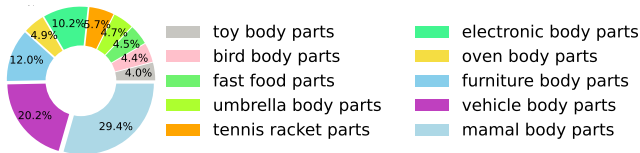

(a) Top 10 missed concepts by ShuffleNetV2.

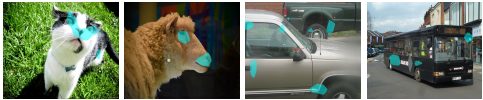

(b) Example images of animals and vehicles.

Figure 9: Semantically categorized explanations of ResNet50-ShuffleNetV2 model pair on COCO dataset.

DELTA's behavior through an overlapping score, computed as the intersection of the regions of *edge* and DELTA divided by the DELTA region. Figure 8 (b) shows example images (with marked activation regions for both *edge* and DELTA networks) for the top two geometrically categorized explanations for AlexNetS's poor performance on CIFAR10. The mean overlapping scores are indicated in parenthesis in the legend for each category. Cyan and magenta patches highlight the regions that DELTA model and *edge* model focus on, respectively. For example, in the first image, AlexNetS fails to classify it as a horse due to a missed eye. However, DELTA precisely captures the eye, correcting the misclassification. DELTA identifies critical features that lower-accuracy *edge* classifiers may overlook but successfully captured by *base* classifiers.

### 6.3 Semantic Categorization

Figure 9 (a) shows the top 10 most frequently missed semantic concepts that the DELTA model brings in to correct ShuffleNetV2's (*edge* model) mispredictions. Notably, about a third of the mispredictions are explained by the model's inability to recognize animal body parts such as eyes, nose, and facial features. Some example images are shown in Figure 9 (b) with activation regions that DELTA captures.

### 6.4 Lost Semantics in Model Compression

We conduct an experiment to understand the semantic loss in a model due to various degrees of model compression. We construct the DELTA model for each pair of *base* model and *edge* models from Table 2. The keywords are extracted from LLM layer to understand the frequency of the missed semantic concepts that corresponds to the loss of accuracy and feature representation capability. Figure 10 shows an example image of a horse correctly predicted by the *base* model but misclassified by the *edge* model. The text output of each model pair explains the difference between the *edge* and *base* models at various levels of detail, which highlights model disparity change after increasing the sparsity rate.

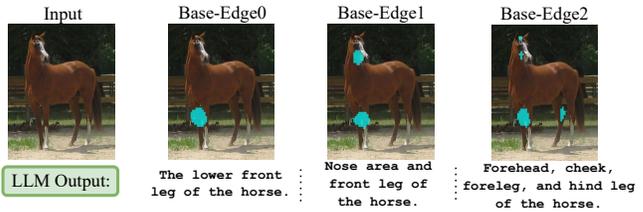

Figure 10: Example LLM output of *base-edge* model pairs.

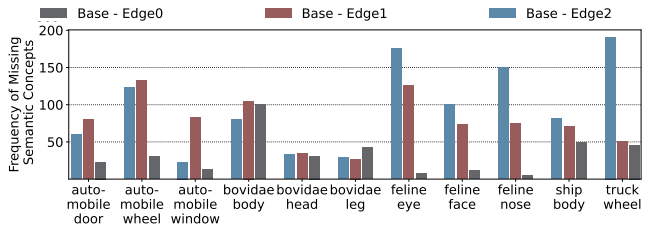

Figure 11: The missing concept frequency comparison between differently compressed *edge* models.

| Models | Param.(M) | FLOPS(M) | Accuracy(%) |
|---|---|---|---|
| **SqueezeNet** | 0.77 | 0.97 | 64.85 |
| **ResNet18** | 11.21 | 2.38 | 81.00 |
| **XDELTA** | 4.55 | 1.98 | 78.86 |

Table 3: Models in XDELTA deployment.

For an aggregate analysis, we identify the top-5 frequently missed concepts for each model pair and create a union list shown in Figure 11 for comparison. As the compression ratio increases, the models lose their ability to capture features in categories like felines, ships, and trucks. However, feature loss for bovidae (mammal with unbranched horns) is less affected due to their distinct recognition features. For complex structures like automobiles, higher compression ratios significantly impact the model's feature capture.

## 7 DEPLOYMENT EXPERIMENT

In order to evaluate the performance of XDELTA in real-world scenarios, we conduct a study involving six participants who contribute a total of 421 pictures of their living and working environments from four different environments. Objects in these images are classified locally on the user's phone using *edge* and DELTA models, and remotely using *base* model, listed in Table 3. XDELTA is applied to summarize the shortcomings of the *edge* model — geometrically as well as semantically. We implement XDELTA for Google Pixel 2XL using pytorch 1.12, which includes *edge* and DELTA models. The *base* and *edge* models are pre-trained on MIT Indoor Scenes dataset of 67 categories. The DELTA model is created and evaluated on the user-contributed data containing eight different categories of images. All images are resized to 256×256 before feeding into the networks.

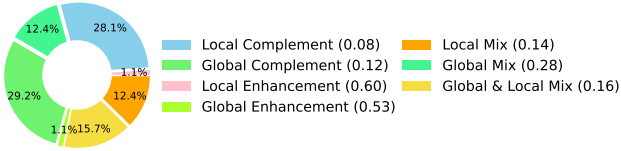

**(a) Summary of geometrically categorized explanations.**

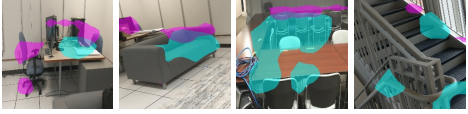

Local Complement      Global Complement

**(b) Examples of local and global feature complement.**

**Figure 12: Geometric explanations on user data.**

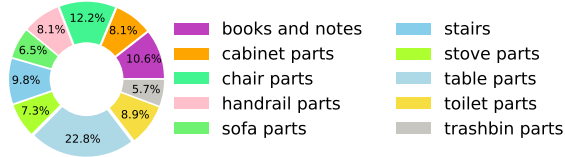

**(a) Top 10 missed concepts by SqeezeNet.**

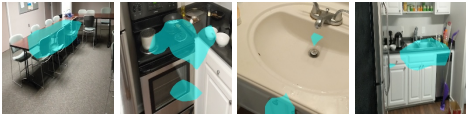

**(b) Examples of meeting room, kitchen and bathroom.**

**Figure 13: Semantic explanations on user data.**

## 7.1 Geometric Categorization

Figure 12 (a) summarizes the cause of incorrect predictions by SqueezeNet (the *edge* model) by geometrically categorizing the explanations. In about 57.3% cases, DELTA brings complementary information from local objects (e.g., sofa seat and table surface) and from spatial contexts (e.g., kitchen area and stairs) as such the *edge* model may focus on certain salient parts with less noisy regions – which is consistent with the dataset-driven experiment for the same pair of models. Figure 12 (b) shows example images (with marked activation regions for both DELTA and *edge* models) for the top two geometrically categorized explanations behind SqeezeNet's poor performance on user-contributed data. As expected, the enhancement category yields higher overlapping scores, while the complement category results in lower scores.

## 7.2 Semantic Categorization

Figure 13 (a) shows the top 10 most frequently missed semantic concepts that the DELTA model brings in to correct SqueezeNet's (*edge* model) mispredictions. For instance, most of the mispredictions are explained by the network's inability to recognize small parts of a larger object such as chair arms and table legs. Some example images (with marked activation regions of DELTA) are shown in Figure 13 (b).

| Datasets | Images | Classes |
|---|---|---|
| CIFAR10 | 60,000 | 10 |
| ImageNet-1K | 1,431,167 | 1,000 |

**(a) Datasets**

| Models | Param. (M) | FLOPS (G) |
|---|---|---|
| VGG16 | 15.25 | 0.314 |
| ResNet50 | 25.56 | 4.112 |
| ResNet56 | 0.86 | 0.127 |

**(b) Base Models**

| Models | Param. (M) | FLOPS (G) |
|---|---|---|
| VGG8 | 4.44 | 0.068 |
| MobileNetV2S | 0.41 | 0.017 |
| SqueezeNet | 1.25 | 0.819 |
| ShuffleNetV2 | 1.37 | 0.043 |
| ResNet8 | 0.08 | 0.013 |
| AlexNet | 61.10 | 0.714 |
| AlexNetS | 23.49 | 0.045 |

**(c) Edge Models**

**Table 4: Models and Datasets**

## 8 COMPONENT ANALYSIS AND ABLATION STUDIES

We conduct a series of experiments to evaluate the effect of individual algorithmic components of the whole framework.

### 8.1 Experimental Setup

**Datasets and Models.** We conduct experiments on two image datasets: CIFAR10 [19] and ImageNet-1K [43]. We use three high-accuracy models (VGG16 [46], ResNet56 [12] and ResNet50 [12]) as the *base* models, which are also used for subgraph extraction method evaluation based on their popularity in the literature. We use seven popular and relatively low-accuracy models as the *edge* models.

Table 4 lists the datasets and models used in this section. We use publicly available pre-trained models [1, 3] whenever possible and train four networks on CIFAR10 by ourselves: VGG8 [46], ResNet8 [12], AlexNetS [20] and MobileNetV2S [44] for 100 epochs with an exponentially decaying learning rate of 0.01. AlexNetS and MobileNetV2S are down-scaled AlexNet [20] and MobileNetV2 [44] to match the CIFAR10 dataset, respectively; Batch normalization layer is not included in VGG8; MobileNetV2S uses multiplier parameter of 0.35; ShuffleNetV2 [32] has 0.5× output channels; SqueezeNet [15] is the 1.0 version. AlexNetS and AlexNet are defined as *edge* models due to their low computational cost and small feature extractor size. All the FLOPS of each model is measured using [2]

**Configurations.** The subgraphs of VGG16 and ResNet56 are extracted on an NVIDIA Tesla K80 GPU. ResNet50 required 16 NVIDIA Tesla V100-SXM2 GPUs due to large-sized model and dataset. The subgraphs are extracted under the magnitude criterion and fine-tuned for 300 epochs. A stochastic gradient descent optimizer with exponential decay and 0.001 learning rate is used in sparsity optimization, $\alpha = 10^{-5}$, and $\beta = 10^{-7}$. We set $\lambda_0 = 0.5$ and $\lambda_1 = 1.0$ in the loss terms. The predefined sparsity rates are $\{\varsigma_i\} \subseteq \{0.125 \times k \mid k = 1, 2, \cdots, 7\}$ for convolutional layers, and $\{\varsigma_i\} \subseteq \{0.2 \times k \mid k = 1, 2, 3, 4\}$ for fully connected layers.

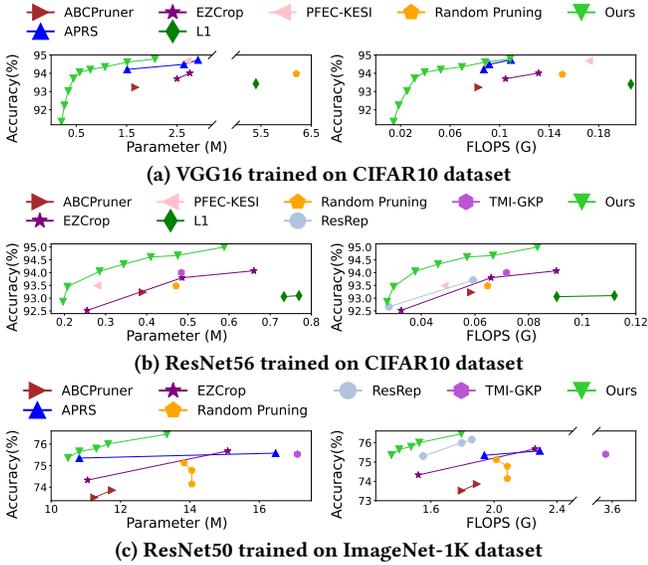

(a) VGG16 trained on CIFAR10 dataset

(b) ResNet56 trained on CIFAR10 dataset

(c) ResNet50 trained on ImageNet-1K dataset

Figure 14: Our subgraph extraction method achieves the best trade-off between accuracy and model compression.

## 8.2 Compactness of DELTA

**Baselines and Metric.** We use structured pruning algorithms that are related to our structured subgraph extraction technique as baselines including several state-of-the-art approaches: $l_1$-norm [22], ABCPruner [26], APRS [49], EZCrop [27], PFEC-KESI [21], ResRep [8], Random Pruning [23], TMI-GKP [57]. We compare the size and FLOPS of the extracted subgraph, and corresponding inference accuracy with our approach. The values shown in Figure 14 are directly reported from their original published papers. The retention of the last few layers in the subgraph is to ensure a fair comparison with other baselines.

**Parameter and FLOPS Reduction.** Figure 14 (a) shows that our subgraph extraction method achieves the highest accuracy of 94.61% after removing 13.74M parameters and reducing 0.226G FLOPS from original VGG16 model. When a more aggressive subgraph extraction is performed on VGG16 that removes 15.05M parameters, our method still achieves 91.35% accuracy. Figure 14 (b) shows similar results. Our method subtracts 0.45M parameters and reduces 0.07G FLOPS from ResNet56, yet achieves the highest accuracy of 94.61%. To further verify our subgraph extraction method, we use a large-scale dataset – ImageNet-1K that contains over 1.4 million images. Figure 14 (c) shows that ResNet50 reaches the highest accuracy of 75.364% after 15.07M parameters and 2.76G FLOPS reduction.

**Convergence.** One of the advantages of our structured subgraph extraction technique is its ability to converge faster. To demonstrate this, we extract structured subgraph from

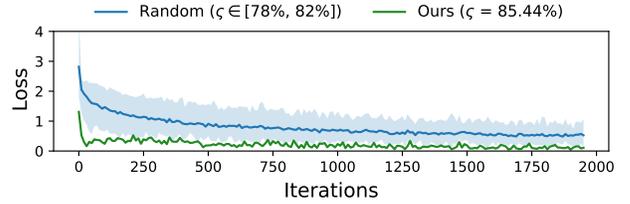

Figure 15: Our subgraph extraction method converges fast and its loss remains lower than the baseline.

| DELTA | Overall Acc. (%) | Δ Acc. (↑ %) | $\mathcal{P}_{D/E}$ | $\mathcal{F}_{D/E}$ | $\mathcal{P}_{D/B}$ | $\mathcal{F}_{D/B}$ |
|---|---|---|---|---|---|---|
| $D_0$ | 92.42 | 8.65 | **0.088** | **0.58** | **0.026** | **0.13** |
| $D_1$ | 93.22 | 9.45 | 0.118 | 0.71 | 0.034 | 0.15 |
| $D_2$ | **93.51** | **9.74** | 0.148 | 0.74 | 0.043 | 0.16 |

(a) Accuracy improvement by DELTA for VGG16-VGG8 pair on CIFAR10. VGG16's accuracy on CIFAR10 is 94.16%.

| DELTA | Overall Acc. (%) | Δ Acc. (↑ %) | $\mathcal{P}_{D/E}$ | $\mathcal{F}_{D/E}$ | $\mathcal{P}_{D/B}$ | $\mathcal{F}_{D/B}$ |
|---|---|---|---|---|---|---|
| $D_0$ | 87.88 | 12.51 | **0.725** | **1.77** | **0.066** | **0.18** |
| $D_1$ | 88.66 | 13.29 | 0.862 | 1.99 | 0.079 | 0.20 |
| $D_2$ | **90.02** | **14.65** | 1.049 | 2.28 | 0.096 | 0.23 |

(b) Accuracy improvement by DELTA for ResNet56-ResNet8 pair on CIFAR10. ResNet56's accuracy on CIFAR10 is 94.37%.

Table 5: Intra-family model pairs.

VGG16 using our method as well as by a baseline strategy that generates and applies random masks from pre-configured sparsity rates for different layers. We use 10 different mask settings. For our method, a single mask is created using averaged sparsity and then applied to a layer. For the baseline, each mask is individually applied to get 10 different sets of results whose mean and variance are used for comparison. Figure 15 shows the cross-entropy loss of both techniques as the model is trained (fine-tuned) on CIFAR10 dataset. We use the same training configurations for fair comparison. We observe that our method not only converges fast but also keeps the loss relatively lower, which ensures its accuracy preservation ability during fine-tuning.

## 8.3 Complementary Capability of DELTA

**Models and Metric.** We take all nine compatible pairs of *base* and *edge* models from Table 4, and generate DELTA models under different parameter and FLOPS constraints. We use the following expressions to express parameter ratio and FLOPS ratio between two models:

$$\mathcal{P}_{D/M} = \frac{param(D)}{param(M)}, \mathcal{F}_{D/M} = \frac{flops(D)}{flops(M)}, M \in \{E, B\}$$

where, $D$, $E$, and $B$ denote DELTA, *edge*, and *base* models, respectively; $param()$ and $flops()$ denote parameter size and FLOPS of the input model. The parameter size and FLOPS of DELTA also include the corresponding values from Feature Resizing Module and Finetuning Module.

**Feature Representation Capability (Intra-family).** Table 5 (a) and Table 5 (b) show the feature representation improvement reflected by improved accuracy due to DELTA for VGG16-VGG8 and ResNet56-ResNet8 pairs, respectively. For each pair, three DELTA models $D_0, D_1, D_2$ are generated by enforcing three different parameter and FLOPS constraints. By extracting less than 2.6%–4.3% parameters from VGG16, DELTA models increase the accuracy of VGG8 by 8.65%–9.74% which is within 0.65%–1.74% of VGG16's accuracy of 94.16%. Although the inclusion of DELTA adds 8.8%–14.8% parameters and 58%–74% FLOPS when compared to the *edge* (VGG8) model, the combined size and FLOPS of *edge* (VGG8) and DELTA model is still 66.57%–68.32% and 62.22%–65.62% less than those of the VGG16 model. We observe a similar trend for the ResNet56-ResNet8 pair. The constraints for the three DELTA models are slightly relaxed considering the large accuracy gap between ResNet56 and ResNet8.

**Feature Representation Capability (Inter-family).** Table 6 (a) and Table 6 (b) show the feature representation improvement reflected by improved accuracy, attributed to DELTA for four model pairs on CIFAR10 and three model pairs on ImageNet-1K, respectively, for different parameter and FLOPS constraints. We observe that by extracting less than 2.47%–32.35% parameters from *base* models on CIFAR10, DELTA models increase the accuracy of *edge* models by 7.61%–17.37% which is within 1.32%–4.50% of ResNet56's accuracy of 94.37% (the highest accuracy). The combined size and FLOPS of *edge* and DELTA model is 21.29%–93.55% and 55.96%–83.17% less than those of the *base* models. We observe a similar trend for the three pairs on ImageNet-1K. In this case, however, the constraints for the DELTA models have been relaxed considering the large accuracy gap between the *base* and the *edge* networks.

**Negative Correlation Evaluation.** We calculate the correlation score [31] between DELTA and *edge* model of the given model pair to evaluate their relationship, which is computed using the samples that are misclassified by *edge* model but correctly predicted by *base* model. The results are shown in Table 7 for each model pair on a representative dataset (CIFAR10). The negative sign denotes a negative correlation between the *edge* model and DELTA, with the correlation score magnitude reflecting the disparity in their feature representation capabilities. Additionally, higher architectural similarity between the *edge* model and DELTA corresponds to a lower correlation score magnitude, indicating reduced complementarity in their represented features.

**Knowledge Distillation (KD) Comparison.** We evaluate the feature representation capability of the *edge* model independently of the DELTA. To maximize the potential representation capability of the *edge* model, we employ the *base*

| Model Pair | Overall Acc. (%) | Δ Acc. (↑ %) | $\mathcal{P}_{D/E}$ | $\mathcal{F}_{D/E}$ | $\mathcal{P}_{D/B}$ | $\mathcal{F}_{D/B}$ |
|---|---|---|---|---|---|---|
| ResNet56 | 89.87 | 7.61 | **0.295** | **1.71** | **0.141** | **0.23** |
| - MobileNetV2S | **93.05** | **10.79** | 0.677 | 2.29 | 0.323 | 0.31 |
| VGG16 | 92.22 | 13.20 | **0.017** | **0.89** | **0.027** | **0.13** |
| - AlexNetS | **92.94** | **13.92** | 0.023 | 1.08 | 0.036 | 0.15 |
| VGG16 | 92.17 | 16.80 | **4.821** | **3.13** | **0.025** | **0.13** |
| - ResNet8 | **92.74** | **17.37** | 6.527 | 3.80 | 0.033 | 0.15 |
| VGG16 | 92.17 | 9.91 | **1.015** | **2.34** | **0.027** | **0.13** |
| - MobileNetV2S | **92.79** | **10.53** | 1.341 | 2.85 | 0.036 | 0.15 |

(a) Accuracy improvement by DELTA for model pairs on CIFAR10. VGG16 and ResNet56 have accuracy of 94.16% and 94.37% on CIFAR10, respectively.

| Model Pair | Overall Acc. (%) | Δ Acc. (↑ %) | $\mathcal{P}_{D/E}$ | $\mathcal{F}_{D/E}$ | $\mathcal{P}_{D/B}$ | $\mathcal{F}_{D/B}$ |
|---|---|---|---|---|---|---|
| ResNet50 | 68.250 | 10.154 | **5.783** | **1.40** | **0.283** | **0.28** |
| - SqueezeNet | **68.870** | **10.774** | 7.018 | 1.59 | 0.343 | 0.32 |
| ResNet50 | 69.018 | 8.366 | **5.193** | **26.90** | **0.278** | **0.28** |
| - ShuffleNetV2 | **70.040** | **9.488** | 5.719 | 30.55 | 0.306 | 0.32 |
| ResNet50 | 68.164 | 11.608 | **0.182** | **1.75** | **0.436** | **0.30** |
| - AlexNet | **68.728** | **12.172** | 0.198 | 1.98 | 0.473 | 0.34 |

(b) Accuracy improvement by DELTA for model pairs on ImageNet-1K. ResNet50's accuracy is 76.146%.

Table 6: Inter-family model pairs.

| Model Pair | VGG16 -VGG8 | VGG16 -ResNet8 | VGG16 -MobileNetV2S | VGG16 -AlexNetS | ResNet56 -ResNet8 | ResNet56 -MobileNetV2S |
|---|---|---|---|---|---|---|
| Correlation Score | -0.0013 | -0.0224 | -0.0584 | -0.0307 | -0.0121 | -0.0149 |

Table 7: Correlation score of different model pairs.

| Base Model (Dataset) | Edge Model | Edge Acc. (%) | Edge + KD Acc. (%) | Edge + DELTA Acc. (%) | $\Delta_{KD}$ Acc. (↑ %) | $\Delta_{E\|\delta}$ Acc. (↑ %) |
|---|---|---|---|---|---|---|
| VGG16 (CIFAR10) | VGG8 | 83.77 | 88.79 | **93.51** | 5.02 | **9.74** |
|  | ResNet8 | 75.37 | 82.20 | **92.74** | 6.83 | **17.37** |
|  | MobileNetV2S | 82.26 | 88.34 | **92.79** | 6.08 | **10.53** |
|  | AlexNetS | 79.02 | 81.67 | **92.94** | 2.65 | **13.92** |
| ResNet56 (CIFAR10) | ResNet8 | 75.37 | 82.31 | **90.02** | 6.94 | **14.65** |
|  | MobileNetV2S | 82.26 | 88.42 | **93.05** | 6.16 | **10.79** |
| ResNet50 (ImageNet-1K) | SqueezeNet | 58.096 | 60.646 | **68.870** | 2.550 | **10.774** |
|  | AlexNet | 56.556 | 60.358 | **68.728** | 3.802 | **12.172** |

Table 8: Accuracy improvement comparison between the distilled *edge* model ($\Delta_{KD}$) and the proposed DELTA approach ($\Delta_{E\|\delta}$).

model as a teacher to transfer distilled knowledge. In Table 8, knowledge distillation improves the *edge* model's accuracy by 2.55%–6.94%, while integrating the DELTA component yields a greater improvement (9.74%–17.37%).

## 9 RELATED WORK

**Model Dissimilarity/Similarity Explanation.** Various approaches [6, 48, 56] of explaining single CNN model's decision making behavior using class activation map [6, 58] or other saliency-map based techniques [48, 56] are proposed in literature, which lacks the capability to understand the model difference. Recent model similarity comparison analysis studies [11, 16, 24] perform measurements across from model functional equivalence to feature-wise distance comparison. However, they fall short of straightforwardly explaining the fine-grained concept-level details of differences between multiple models' decision-making process.

**Structured Pruning.** Structured pruning methods based on property importance [27, 54] limit model compression ratio and accuracy. Adaptive importance based approaches [10, 29] require specific design to monitor the filter importance and carefully tuned hyperparameters. Automatic sparsity search, as presented in the literature [26, 49], faces challenges in finding the optimal substructure due to the extensive search space and the trade-off between searching efficiency and accuracy. Our method utilizes a more compact search space to find the optimal subnetwork without special ranking technique to achieve state-of-the-art performance.

## 10 CONCLUSION

This paper presents XDELTA, a differential explainable AI tool designed to elucidate the distinctions between a lower-accuracy *edge* model and a higher-accuracy *base* model across multiple levels of detail. Central to this approach is the introduction of DELTA, a neural network architecture that captures the differences between model pairs. By augmenting the feature representation capabilities of the *edge* model, DELTA enables the combined system to achieve a feature representation on par with the *base* model. Utilizing the complementarity of DELTA, XDELTA offers explanations at various levels, ranging from high-level geometric insights to fine-grained, human-understandable details.